\documentclass[conference]{IEEEtran}

\IEEEoverridecommandlockouts
\usepackage{times}

\usepackage[numbers]{natbib}
\usepackage{multicol}
\usepackage[bookmarks=true]{hyperref}
\usepackage{amsmath}
\usepackage{amsfonts}
\usepackage{algorithm}
\usepackage{algorithmic}
\usepackage{hyperref}
\usepackage{rotating} 
\usepackage{subcaption}
\usepackage{stfloats}

\begin{document}

\title{Continuous Trajectory Optimization via B-splines for Multi-jointed Robotic Systems}

\author{Changhao Wang*, Ting Xu* and Masayoshi Tomizuka
\thanks{*: Both authors contributed equally to this work.}
\thanks{Department of Mechanical Engineering, University of California, Berkeley, CA, USA.
        {\tt\small \{changhaowang, ting\_xu, tomizuka\}@berkeley.edu}}%
}
\maketitle

\begin{abstract}
Continuous formulations of trajectory planning problems have two main benefits. First, constraints are guaranteed to be satisfied at all times. Secondly, dynamic obstacles can be naturally considered with time.
This paper introduces a novel B-spline based trajectory optimization method for multi-jointed robots that provides a continuous trajectory with guaranteed continuous constraints satisfaction. At the core of this method, B-spline basic operations, like addition, multiplication, and derivative, are rigorously defined and applied for problem formulation. B-spline unique characteristics, such as the convex hull and smooth curves properties, are utilized to reformulate the original continuous optimization problem into a finite-dimensional problem. Collision avoidance with static obstacles is achieved using the signed distance field, while that with dynamic obstacles is accomplished via constructing time-varying separating hyperplanes. Simulation results on various robots validate the effectiveness of the algorithm. In addition, this paper provides experimental validations with a 6-link FANUC robot avoiding static and moving obstacles.
\end{abstract}

\IEEEpeerreviewmaketitle

\section{Introduction}
Trajectory optimization is to find motions that satisfy movement constraints and optimize some aspects of the movement. Since a trajectory is continuous, the trajectory planning problems should be formulated as continuous optimization problems with infinite degrees of freedom. 


To solve the continuous problem, researchers have proposed different methods to approximate the solution. Most optimization-based methods~\cite{ratliff2009chomp, kalakrishnan2011stomp, park2012itomp, wang2022bpomp, wang2021trajectory} discretize the continuous trajectory into waypoints and utilize numerical optimization methods to solve the problem. This strategy has been successfully applied to many problems in robotics navigation~\cite{ratliff2009chomp, kalakrishnan2011stomp, schulman2013finding}, manipulationn~\cite{mason2018toward, jin2019robust, tang2018framework, jin2021trajectory, wang2022offline}, and control~\cite{wang2022safe, sun2021online}. However, the discretized solution may not guarantee constraint satisfaction for the continuous problem. This kind of method is likely to obtain infeasible solutions. In addition, discretized methods generally find it hard to deal with dynamic obstacles. Methods like ITOMP~\cite{park2012itomp} require that trajectories of the robot and obstacle do not intersect within a whole sampling interval, while continuous formulations only need to be constrained on the specific time.

TrajOpt~\cite{schulman2013finding} considers collisions of continuous trajectories to some extent via a convex hull approximation. However, the convex hull approximation introduced in TrajOpt~\cite{schulman2013finding} has no rigorous guarantee when rotational trajectories are involved, especially in high dimensional spaces. BPOMP~\cite{wang2022bpomp} proposes to add a constraint on the `worst state' and formulate the problem as a bilevel optimization to handle the collisions that happen in the middle of the waypoints. However, due to the non-convexity of the problem, solving that bilevel problem is not trivial, and its relaxation to a standard NLP may be overly conservative. 

Another way to deal with the continuous problem is via function parameterization to reformulate the problem in the parameter space. Chen~\cite{chen1991solving} parameterized optimal trajectories by B-splines and relaxed the problem based on the convex hull property. Van Loock et al.\cite{van2015b} validated that the convex hull relaxation of B-splines can guarantee constraint satisfaction at all instances. The conservatism introduced from the relaxation is also discussed. Mercy et al.~\cite{mercy2016real}~\cite{mercy2018spline} \cite{mercy2017spline} proposed a B-spline based continuous collision avoidance formulation by constructing time-varying separating hyperplanes. This formulation takes advantage of trajectory timing, and it's able to obtain solutions with minimal conservatism. They validated the continuous safety of the solution on mobile robots and CNC machines. Yet, their continuous B-spline formulation cannot generalize directly to multi-jointed robots, and the obstacles are restricted to simple geometries such as rectangles and circles. Computation time also scales up as the number of obstacles increases due to the use of the separating hyperplane for each obstacle. 
\textbf{\begin{figure}[t]
\centerline{\includegraphics[scale = 0.33]{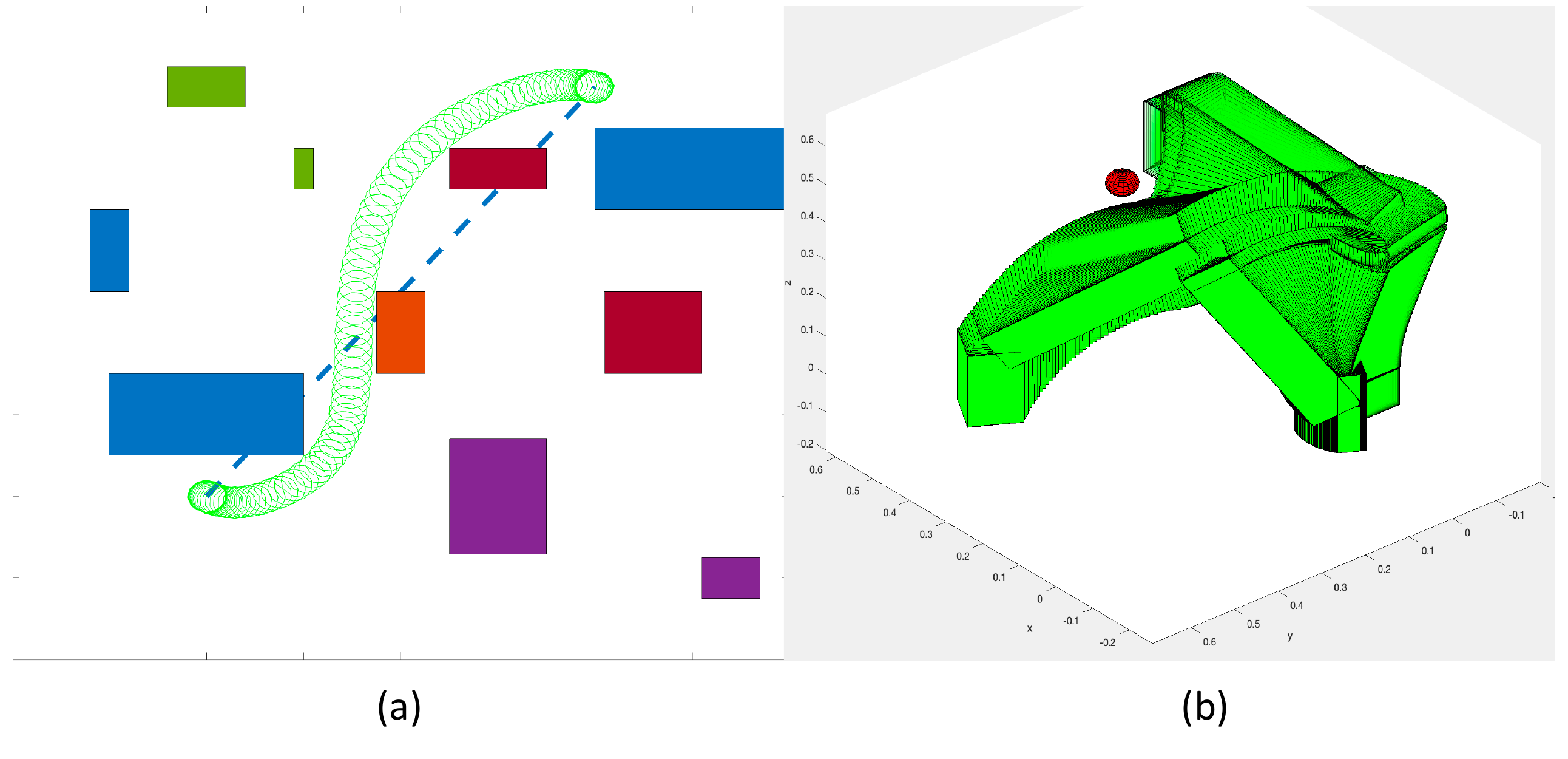}}
\caption{Two simulation results of the proposed B-spline based continuous trajectory optimization method. (a) shows a mobile robot navigating through several obstacles in 2D. (b) shows a six-link robot avoiding a spherical obstacle.
}
\label{fig_Cover}
\end{figure}}

The main contribution of this paper is the extension of the continuous formulation in~\cite{mercy2017spline} to multi-jointed robots with high-DOF by formulating the dynamics and forward kinematics via B-splines. In addition, the efficiency for collision checking is improved by applying the signed distance field for static obstacles while still utilizing the time-varying separating hyperplanes for dynamic obstacles.
To achieve the continuous formulation for multi-jointed robots, we view each joint as an individual robot with additional dynamic and kinematic constraints. B-spline basic operations, like addition, multiplication, and derivative, are defined to formulate the continuous problem. In the end, the convex hull property of B-splines is applied to relax the infinite-dimensional optimization problem. It is worth noting that applying the convex hull property of B-splines for problem relaxation differs from the convex hull approximation introduced in TrajOpt~\cite{schulman2013finding} or ITOMP~\cite{park2012itomp}, where the convex hull approximation of B-splines has a guarantee of constraints satisfaction no matter what the trajectories are. Several simulation and experimental results validate the effectiveness of the proposed method.


The remaining paper is organized as follows. Section II describes B-spline properties and basic operations. We present our continuous B-spline based trajectory planning method for multi-jointed robots in Section III. Section IV tests the performance of the proposed method through a series of simulations and experiments. 
In the end, Section V concludes the paper.

\section{B-spline Properties and Operations}

\subsection{B-spline Definition}
B-splines are linear combinations of basis functions $S(t) = \sum_{i=0}^{n} c_i B_{i,p}(t)$, where $c_i$ are the control points (coefficients), $B_{i,p}$ are the basis functions, $n$ is the number of control points, and $p$ is the spline order (degree)~\cite{piegl2012nurbs}. Basis functions are defined by the spline order $p$,  time $t$, and a knot vector $U = [u_0,u_1, \cdots, u_m]$. The relation between these three parameters is $n = m - p - 1$.

The basis function of B-splines can be defined recursively according to De Boor's algorithm~\cite{de1977package}. The $0^{th}$ order basis function $B_{i,0}$ is defined between knot intervals:
\begin{equation}
    B_{i,0}(t)= 
\begin{cases}
    1& \text{if } u_i \le t < u_{i+1}\\
    0 & \text{otherwise}
\end{cases}
\end{equation}
The higher order basis function $B_{i,p}$ is defined recursively as:
\begin{equation}
    B_{i,p}(t) = \frac{u - u_i}{u_{i+p} - u_i} B_{i, p-1}(t) + \frac{u_{i+p+1} - u}{u_{i+p+1} - u_{i+1}} B_{i+1, p-1}(t)
\label{eqn_cox_de_Boor}
\end{equation}


\subsection{B-spline Properties}
\begin{figure}
    \centering
    \includegraphics[scale=0.3]{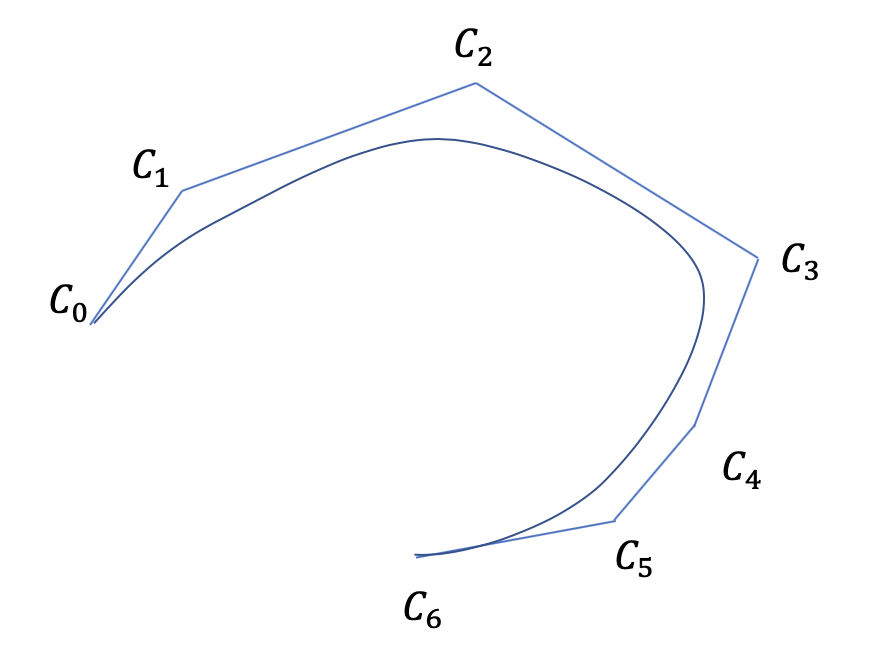}
    \caption{An example of a clamped B-spline and its convex hull property.}
    \label{fig:clamped}
  \end{figure}

Taking the advantage of the recursive formulation, B-splines have several useful properties. We list five of them that are used in the proposed method.
\begin{enumerate}
  \item Any B-spline basis function $B_{i,p}(t)$ is non-negative
  \item The summation of all basis functions of order $p$ at any time $t$ is equally to one: $\sum_{i=0}^{n} B_{i,p}(t) = 1$. 
  \item The derivative of B-splines are still B-splines
  \item Clamped endpoints: B-splines generally do not pass through starting and end control points. Repeating the first and last knots $p$ time can force the B-spline starts and ends at specified control points, which would return a clamped B-spline as shown in Fig. \ref{fig:clamped}
  \item Strong convex hull property: Any B-spline is contained within a convex hull formed by its control points. Mathematically, this property extends to a relaxation that $a < c_i  < b$ is a sufficient condition of $a < S(t) = \sum_{i=1}^{n} c_i B_{i,p}(t) < b$.
\end{enumerate}

\subsection{B-spline Operations}
To formulate the problem with B-splines, we need to define function operations. Basic operations for B-splines, such as addition, multiplication, and derivative, are not as easy as that for scalar functions, especially between two B-splines that have different orders and knot vectors. To properly define a B-spline, we need to know both basis functions, which are determined by the spline order and the knot vector, and the coefficient of each basis function. There are some papers studied on B-spline operations \cite{ueda1994multiplication}, and we utilize a simplified version of the algorithm proposed in \cite{van2016omg} for better efficiency. In general, a curve can have infinite B-spline representations, which means it can have different basis functions and coefficients. In this paper, we follow a strategy that calculates a feasible basis function and then solves a least square problem to obtain the set of coefficients.

For B-spline addition, we consider the most general case that two B-splines $S_1(t) = \sum_{i=0}^{n_1} c_{1i} B_{i,p_1}(t)$, $S_2(t) = \sum_{i=0}^{n_2} c_{2i} B_{i,p_2}(t)$ have different orders $p_1$, $p_2$ and different knot vectors $U_1$, $U_2$. The union of $U_1$ and $U_2$ is a choice of the knot vector of the resulting B-spline. Since addition will not increase the function order, the order of the resulting B-spline should be equal to the maximum order of $S_1(t)$ and $S_2(t)$. Knowing the knot vector and the order is sufficient to calculate basis functions according to (\ref{eqn_cox_de_Boor}). To find the coefficients of the basis function, we formulate the problem as an optimization problem shown in (\ref{eqn_spline_transformation}), where $f(\cdot,\cdot)$ is either addition or multiplication of two B-splines. A necessary optimality condition for (\ref{eqn_spline_transformation}) is that the value should be equal at specific time instances. Therefore, we approximate the solution by discretizing the problem and solving it via least squares. We follow a similar procedure for the multiplication of two B-splines. The only difference is the 
spline order of the resulting B-spline should be $p_3 = p_1 + p_2$. The details of the B-spline addition and multiplication is shown in Algorithm \ref{alg_operation}.

\begin{equation}
\begin{aligned}
\min_{c_3} \quad & \|f(S_1(t),S_2(t)) - \sum_{i=0}^{n_3} c_{3i} B_{i,p_3}(t)\|^2\\
\textrm{s.t.} \quad & 
S_1(t) = \sum_{i=0}^{n_1} c_{1i} B_{i,p_1}(t)\\
& S_2(t) = \sum_{i=0}^{n_2} c_{2i} B_{i,p_2}(t)\\
\end{aligned}
\label{eqn_spline_transformation}
\end{equation}

\begin{algorithm}
\caption{B-spline addition and multiplication}
\begin{algorithmic}[1] 
\label{alg_operation}
\REQUIRE $S_1$, $U_1$, $p_1$, $S_2$, $U_2$, $p_2$
\IF{addition}
    \STATE $p_3 = max(p_1,p_2)$ 
\ELSIF{multiplication}
    \STATE $p_3=p_1+p_2$
\ENDIF
\STATE $U_3$ = $U_1 \cup U_2$
\STATE $B_{i,p_3}\leftarrow$ De Boor's algorithm(\ref{eqn_cox_de_Boor}) 
\STATE $c_3\leftarrow$ Solve (\ref{eqn_spline_transformation})
\RETURN $S_3(t) = \sum_{i=0}^{n_3} c_{3i} B_{i,p_3}(t)$
\end{algorithmic}
\end{algorithm}

According to the B-spline property (3), the derivative of a B-spline can still be represented by a B-spline. Suppose a B-spline $S(t) = \sum_{i=0}^{n} c_i B_{i,p}(t)$ with degree $p$ and knot vector $U$, and its derivative is denoted as $\dot{S}(t) = \sum_{i=0}^{n-1} d_i \dot{B}_{i,p-1}(t)$ with the knot vector $\dot{U}$. The relationship between $S(t)$ and $\dot{S}(t)$ is shown in (\ref{eqn_derivative}). The same equations and rules apply for taking higher order derivatives recursively~\cite{piegl2012nurbs}.
\begin{equation}
\begin{aligned}
    \dot{U} &= U[1:end-1]\\
    d_i &= \frac{c}{u_{i+p+1} - u_{i+1}} (c_{i+1} - c_i)\\
    \dot{B}_{i,p-1} &= \frac{p}{u_{i+p} - u_i} B_{i,p-1} - \frac{p}{u_{i+p+1} - u_{i+1}} B_{i+1,p-1}  
\end{aligned}
\label{eqn_derivative}
\end{equation}

\section{B-Spline Based Continuous Trajectory Optimization}
In this section, we present the B-spline based continuous trajectory optimization formulation for multi-jointed robots. The trajectory of a multi-jointed robot can be represented by a function in the joint space. We decouple the trajectory planning problem for a multi-jointed robot to problems of each link of the robot, and the single link motion in the Cartesian space is obtained by the forward kinematics. This section is organized as follows. First, we discuss the general problem formulation for a single-link mobile robot. Secondly, we introduce the continuous formulation of forward kinematics and general dynamic constraints. In the end, the collision avoidance constraint formulation will be discussed.
\subsection{Problem Formulation}
Planning an optimal collision-free trajectory for mobile robots can be formulated as an optimization problem. The objective could either be minimizing the overall execution time, the actual distance traveled by the robot, or other metrics that users are interested in. Constraints should include initial and final conditions in positions, velocities, and accelerations, as well as their ranges. Collision, kinematic, and dynamic constraints should also be taken into consideration. The general formulation~\cite{mercy2016real} of the trajectory optimization problems along with all constraints can be listed as follows,
\begin{equation}
\begin{aligned}
\min_{q(\cdot),T} \quad & T\\
\textrm{s.t.} \quad & 
q(0) = q_{initial}, \ \ q(T) = q_{goal}\\
&\dot{q}(0) = 0, \ \ \dot{q}(T) = 0\\
&\ddot{q}(0) = 0, \ \ \ddot{q}(T) = 0\\
&\dot{q}_{min} \leq \dot{q}(t) \leq \dot{q}_{max}\\ 
&\ddot{q}_{min} \leq \ddot{q}(t) \leq \ddot{q}_{max}\\
& \text{obstacle avoidance}\\
& \text{kinematics, dynamics constraints}
\end{aligned}
\label{eqn_ori_formulation}
\end{equation}
where $q(\cdot)$ denotes the continuous trajectory of the robot, and $T$ represents the overall execution time. The proposed method utilizes B-splines for trajectory parameterization. Take a mobile robot traveling in a 3D environment as an example:
\begin{equation}
    q(\tau) = \begin{bmatrix}
   q_x(\tau) \\
   q_y(\tau) \\
   q_z(\tau)
\end{bmatrix} = \sum_{i=0}^{n} 
\begin{bmatrix}
   c_{xi} \\
   c_{yi} \\
   c_{zi}
   \end{bmatrix} B_{i,p}(\tau)
\label{eqn_vehicle_trajectory}
\end{equation}
where time is used in the dimensionless form in order to relate $T$ in constraint formulations:
\begin{equation}
    \tau = \frac{t}{T}
\end{equation}
and $T$ is the total travel time. To simplify the formulation in constraints, a new matrix $C \in \mathbb{R}^{n\times 3}$ is created to collect all coefficients:
\begin{equation}
    C = [c_x \ c_y \ c_z]
\end{equation}

With B-spline representations established, the method then proceeds to formulate the objective and constraints. With the convex hull property of B-splines, the state variables are now replaced with B-spline coefficients. This relaxation applies to constraints as well. 
\begin{equation}
\begin{aligned}
\quad & 
C(1,:) = q_{initial}, \ \ C(end,:) = q_{goal}\\
&d{C}(1,:) = 0, \ \ d{C}(end,:) = 0\\
&d^2{C}(1,:) = 0, \ \ d^2{C}(end,:) = 0\\
&\dot{q}_{min}T \leq d{C}(i,:) \leq \dot{q}_{max}T\\
&\ddot{q}_{min}T^2 \leq d^2{C}(i,:)\leq \ddot{q}_{max}T^2\\
& \text{obstacle \ avoidance}\\
& \text{kinematics, dynamics constraints}
\end{aligned}
\label{eqn_relax_constraints}
\end{equation}
where $dC = [dc_x \ dc_y \ dc_z]$ denotes the coefficients of the derivative of B-spline $\dot{q}(\cdot)$, which can be obtained in a closed form by applying (\ref{eqn_derivative}).

\subsection{Forward Kinematics and Dynamics for Multi-jointed Robots}

For a multi-jointed robot, trajectories are represented in the joint space. We view the trajectory planning problem for a multi-jointed robot as a problem for each link of the robot. The motion of the link is determined by applying forward kinematics to transform the trajectory from the joint space to the Cartesian space. In this subsection, we introduce the B-spline continuous formulation of the forward kinematics transformation and discuss general dynamic constraints. 

According to \cite{siciliano2010robotics}, the Denavit-Hartenberg parameters of the $i$th link are denoted by $\alpha_i$, $\theta_i$, $a_i$, and $d_i$, respectively. 
The homogeneous transformation matrix from the $(i-1)$th coordinate to the $i$th coordinate can be expressed by $T_i^{i-1}$ shown in (\ref{eqn_homogenous}).
\begin{equation}
T_i^{i-1} = \begin{bmatrix}
   R_i^{i-1} & p^i \\
    \Vec{0} & 1
\end{bmatrix}
\label{eqn_homogenous}
\end{equation}

\begin{equation}
R_i^{i-1} = \begin{bmatrix}
    \cos(\theta_i) & -\sin(\theta_i)\cos(\alpha_i) & \sin(\theta_i)\sin(\alpha_i) \\
    \sin(\theta_i) & \cos(\theta_i)\cos(\alpha_i) & -\cos(\theta_i)\sin(\alpha_i) \\
    0 & \sin(\alpha_i) & \cos(\alpha_i) \\
\end{bmatrix}
\end{equation}

\begin{equation}
p^i = \begin{bmatrix}
    a_i\cos(\theta_i) \\
    a_i\sin(\theta_i) \\
    d_i \\
\end{bmatrix}
\end{equation}
For a prismatic joint, $d_i$ is the joint space variable, and the homogeneous transformation matrix $T_i^{i-1}$ can fit the proposed B-spline formulation by substituting $d_i$ with a B-spline. However, the homogeneous transformation matrix contains trigonometric operations on the joint variable $\theta_i$ for a revolute joint. The trigonometric operations on B-spline functions are not clear. 

To deal with this problem, instead of directly establishing B-spline parameterization on $\theta_i$, a change of variable to $q_i$ shown in (\ref{eqn_change_variable}) is able to transform the trigonometric operations to B-spline basic operations. $\cos(\theta_i)$ and $\sin(\theta_i)$ can then be written in terms of $q_i$ by B-spline basic operations (\ref{eqn_cos_sin}).
\begin{equation}
    q_i (\tau ) = \tan(\frac{\theta_i (\tau)}{2}) = \sum_{j=0}^{n} c_{i_{j}} B_{j,3}(\tau)
\label{eqn_change_variable}
\end{equation}

\begin{equation}
\begin{aligned}
    \cos(\theta_i(\tau)) &= \frac{1-q_i(\tau)^2}{1+q_i(\tau)^2}\\
    \sin(\theta_i(\tau)) &= \frac{2q_i(\tau)}{1+q_i(\tau)^2}
\end{aligned}
\label{eqn_cos_sin}
\end{equation}

As a result, the transformation matrix with the changed variable is shown in (\ref{eqn_Transformation}). Since $1+q_i(\tau)^2$ is always greater than $0$, we can always multiply the denominator on both sides of the constraint and make it only consists of B-spline multiplications and additions. 

\begin{figure*}[b]
\normalsize
\hrulefill
\begin{equation}
T_i^{i-1} = \frac{1}{1+q_i(\tau)^2} \begin{bmatrix}
    1-q_i(\tau)^2& -(1+q_i(\tau)^2)\cos(\alpha_i) & 2q_i(\tau)\sin(\alpha_i) & a_i(1-q_i(\tau)^2) \\
    2q_i(\tau) & (1-q_i(\tau)^2)\cos(\alpha_i) & -(1-q_i(\tau)^2)\sin(\alpha_i) & 2a_iq_i(\tau) \\
    0 & \sin(\alpha_i)(1+q_i(\tau)^2) & \cos(\alpha_i)(1+q_i(\tau)^2) & d_i(1+q_i(\tau)^2) \\
    0 & 0& 0 & 1+q_i(\tau)^2
\end{bmatrix}
\label{eqn_Transformation}
\end{equation}
\end{figure*}

Conversely, after we obtained the optimal value of $q_i$, we can calculate $\theta_i$ by applying the inverse tangent function $\theta = 2\cdot \tan^{-1}{q}$. This change of variable was introduced by Mercy~\cite{mercy2017spline} to deal with trigonometric operations that appeared in nonholonomic vehicles where rotations are within the range of $[-\pi,\pi]$. However, the range of joint angles in many industrial robots may exceed the domain of the definition of the inverse tangent function. Therefore, the change of variable (\ref{eqn_change_variable}) may not be able to work. We extend this idea to more general cases by making a modification:
\begin{equation}
    q_i (\tau ) = \tan(\frac{\theta_i (\tau)}{2^n}), \ n\in\mathbb{Z}^+
\label{eqn_change_variable_new}
\end{equation}
By applying the new change of variable, the maximal range of the joint angle $\theta$ is expanded to $[-2^{n-1}\pi,2^{n-1}\pi]$, and trigonometric operations can still be similarly transformed to B-spline basic operations by recursively applying the tangent half-angle substitution. For simplicity of the expression, we choose $n=1$ for later derivation, however, it should be easy to generalize to any choice of $n \in \mathbb{Z}^+$.

\begin{figure}[t]
\centerline{\includegraphics[scale = 0.8]{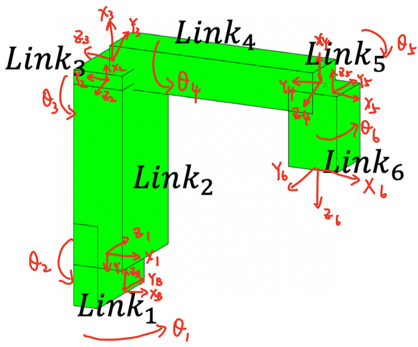}}
\caption{Illustration of the forward kinematics and link vertices of a 6-link robot
}
\label{fig_local_frame}
\end{figure}

Shown in Fig. \ref{fig_local_frame}, a 6-link robot is approximated by six cuboids, and the local frame for each joint is defined according to \cite{siciliano2010robotics}. Forward kinematics is defined as follows:
\begin{equation}
    T_0^i(\tau) = FK_i(q(\tau))=T_0(\tau)\prod_{j=1}^i T_j^{j-1}(\tau)
\label{eqn_fk}
\end{equation}
where $T_0$ denotes the base frame and $FK_i(\cdot)$ represents the forward kinematic transformation for the $i$th link. By applying forward kinematics, we can transform any points $p$ in the local frame $p^i$ to the point in the base frame $p^0$:
\begin{equation}
    p^0(\tau) = FK_i(q(\tau))p^i(\tau)
\label{eqn_change_frame}
\end{equation}
Utilizing (\ref{eqn_change_frame}), we are able to transform the joint space states of multi-jointed robots to the position of each link in the Cartesian space. Therefore, the trajectory planning problem for a multi-jointed robot can be decoupled into several problems for each link with additional forward kinematics constraints (\ref{eqn_change_frame}).

Dynamics models are required for optimal motor torque planning of the robot. The proposed B-spline based trajectory optimization method is able to directly take continuous dynamics models into the optimization formulation without any discretization.

The dynamics model of a robot can be characterized by an ODE function $\dot{q} = f(q)$, where $q$ represents the state of the robot, such as joint angles and velocities, and $f(\cdot)$ denotes the model. From B-spline property (3), if $q$ is represented by a B-spline, the derivative of $q$ is still a B-spline, and its coefficients can be expressed in a closed form as shown in (\ref{eqn_derivative}). In the end, the dynamics model can be relaxed and transformed into an equation that only relates to coefficients of $q$ via the convex hull property.

\subsection{Collision Avoidance}

Shown in Fig. \ref{fig_local_frame}, a multi-jointed robot is approximated by several cuboids for collision checking. To ensure the safety of the trajectory, we check the collision for each of the links to guarantee the whole robot is collision-free.

Static and dynamic obstacles are involved in collision avoidance constraints. Since the static obstacles do not move throughout time, it is common and straightforward to pre-compute a signed distance field. Signed-distance field can be viewed as a lookup chart that stores the distance between every grid in the workspace to its nearest obstacle boundary~\cite{ratliff2009chomp}. The distance value is negative inside obstacles, positive outside, and zero at the boundary. Therefore, the collision avoidance constraint is equivalent to making every point on the robot have a positive signed distance value throughout the time. The constraint is formulated in (\ref{eqn_signeddistance}), where $p^i$ denotes any point of the link in the $i$th frame, and $SD(\cdot)$ is the signed distance function.


\begin{equation}
    SD(FK_i(q(\tau))p^i) \geq 0
\label{eqn_signeddistance}
\end{equation}

By applying the convex hull property of the B-spline, we can relax the inequalities shown in (\ref{eqn_signeddistance}) to inequalities only on the coefficients of the B-spline.


However, it's not efficient to update the entire signed distance field for moving obstacles, instead, we utilize the formulation introduced in \cite{mercy2016real} by constructing time-varying separating hyperplanes.

The basic idea of the separating hyperplane theorem is that, if two convex objects are not in collision, there should exist a hyperplane that separates the two objects. Translated into constraints, for a cuboid link and a spherical obstacle with radius $d_o$, this gives: 
\begin{equation}
\begin{aligned}
    a(\tau)\cdot q_{rj}(\tau) + b(\tau) &\geq 0,\ j= 1,\cdots 8 \\
    a(\tau)\cdot q_o(\tau) + b(\tau) +d_o &\leq 0\\
    \|a(\tau)\|_2^2-1 &\leq 0
\end{aligned}
\label{eqn_separatinghyperplane}
\end{equation}
where $q_{rj}(\tau)$ denotes the trajectory of the $j$th vertex of the link and $q_{o}(\tau)$ represents the obstacle trajectory. $a(\tau)$ is the normal vector to the hyperplane and $b(\tau)$ is hyperplane’s offset. (\ref{eqn_separatinghyperplane}) is also generalizable for different convex object shapes, and different obstacle trajectories. 

This continuous formulation naturally considers the dynamic obstacles with time. Therefore, no conservative approximations are needed as introduced in ITOMP~\cite{park2012itomp}. Compared to other collision-checking methods that utilize the separating hyperplane theorem, this formulation does not require to specifically calculate of the parameters of the hyperplane $a(\tau)$ and $b(\tau)$ in advance. Since the existence of the hyperplane is sufficient to guarantee the safety of the trajectory, we just add $a(\tau)$ and $b(\tau)$ to our optimization variables and let the solver find a feasible value. If $q_{rj}(\tau)$, $q_o(\tau)$, $a(\tau)$, and $b(\tau)$ in (\ref{eqn_separatinghyperplane}) are represented by B-splines, we further simplify the constraints. The left-hand side of each inequality constraint can be transformed to a single B-spline according to B-spline operations, and with the convex hull property of B-splines, we are able to relax these constraints to finite dimensions. 

All in all, the collision avoidance constraints are formulated in (\ref{eqn_signeddistance}) for static obstacles, and (\ref{eqn_separatinghyperplane}) for dynamic obstacles. B-spline operations and properties can be used to further relax the continuous constraints. The computation time of the signed distance formulation is not sensitive to the number of obstacles, and the signed distance field is able to represent complicated non-convex obstacles. The combination of the signed distance field and time-varying separating hyperplanes is able to achieve efficient and effective performance for continuous collision checking.

\section{Simulation and Experimental Results}

In this section, the comparison of computation time between the signed distance field and the separating hyperplane formulation for mobile robots is shown to validate the effectiveness of combining them together in terms of efficiency and performance. Simulations and experimental results on multi-jointed validate the effectiveness of the method. 

\subsection{Signed Distance Field vs Separating Hyperplanes}

The proposed method utilizes signed distance fields for collision checking in static environments and separating hyperplanes for dynamic obstacles, whereas Mercy et al.~\cite{mercy2016real} proposed to use separating hyperplanes for all. While constructing separating hyperplanes is a great way for taking continuous trajectory collision into consideration, it takes time to compute. In this section, we show simulations that the use of signed distance for static obstacles contributes to saving computation time, while still ensuring continuous safety.

\begin{figure}[t]
\centerline{\includegraphics[scale = 0.67]{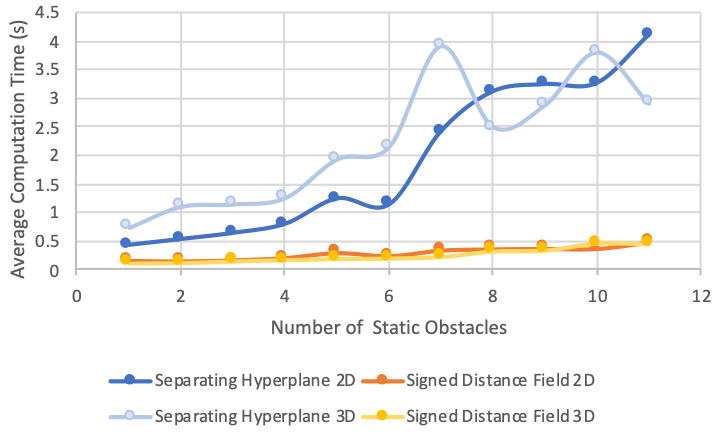}}
\caption{Comparison between the computation time of the signed distance and separating hyperplane formulations for collision avoidance constraints. The figure shows that the signed distance formulation increase is much more efficient than constructing separating hyperplanes when the number of static obstacles is large.
}
\label{fig_compare}
\end{figure}

We tested the proposed method with a spherical holonomic vehicle with radius $d_r$ in both 2D and 3D simulated in MATLAB 2019b. CasADi~\cite{andersson2012casadi} with ipopt non-linear solver is used as the optimization solver. All computations are done in MATLAB 2019b with a 2.7 GHz quad-core Intel i7 processor. 
The trajectory of the vehicle $q(\tau)$ is represented by a clamped B-spline shown in (\ref{eqn_vehicle_trajectory}) with an order equal to $3$. 
The average computation time of the proposed method for mobile robots is shown in Fig. \ref{fig_compare}. It is clear that the computation time is increasing with the increase of obstacles if constructing a separating hyperplane for each of them, where the computation time for the signed distance field remains the same. More specifically, the computation time of the signed distance field formulation is more than $8$ times faster than the separating hyperplane formulation when the number of static obstacles is greater than $10$.
Therefore, it shows that the combination of the separating hyperplane and signed distance field formulation for collision avoidance is able to make the algorithm efficient when the number of static obstacles is large. 


\subsection{Simulation Results on Multi-jointed Robots}

We tested the proposed method on various robots with revolute joints. We set the state of multi-jointed robots as $q_i (\tau ) = \tan(\frac{\theta_i (\tau)}{2n})$ as we discussed in section III.B. Similar to the case of the mobile robot, the state of the robot $q$ is represented by a B-spline, and the optimization formulation (\ref{eqn_ori_formulation}) is simplified by B-spline properties and operations. The difference is that the multi-jointed robots require forward kinematics to transform the joint space angles to poses in the Cartesian space, and we need to make sure all the links of the robot are collision-free.

The proposed method for multi-jointed robots was first implemented on a three-link robot for validation. By fixing the base, only rotation around the z-axis is allowed for the first link, and pitch around the y-axis is allowed for the second and third links. The Denavit-Hartenberg kinematic parameters are set as:
\begin{center}
\label{table2}
\begin{tabular}{|c| c| c| c| c|} 
\hline
Link & $a_i$(m)  & $\alpha_i$(rad) & $d_i$(m) &  $\theta_i$(rad) \\ \hline
1    & $0.5$& $-\frac{\pi}{2}$ & $0$ & $\theta_1$\\ \hline
2   & $0.44$  & $\pi$ & $0$ & $\theta_2$\\ \hline
3   & $0.35$ & $-\frac{\pi}{2}$ & $0$ & $\theta_3$\\ \hline
\end{tabular} 
\end{center}

\begin{figure}
    \centering
    \includegraphics[scale=0.7]{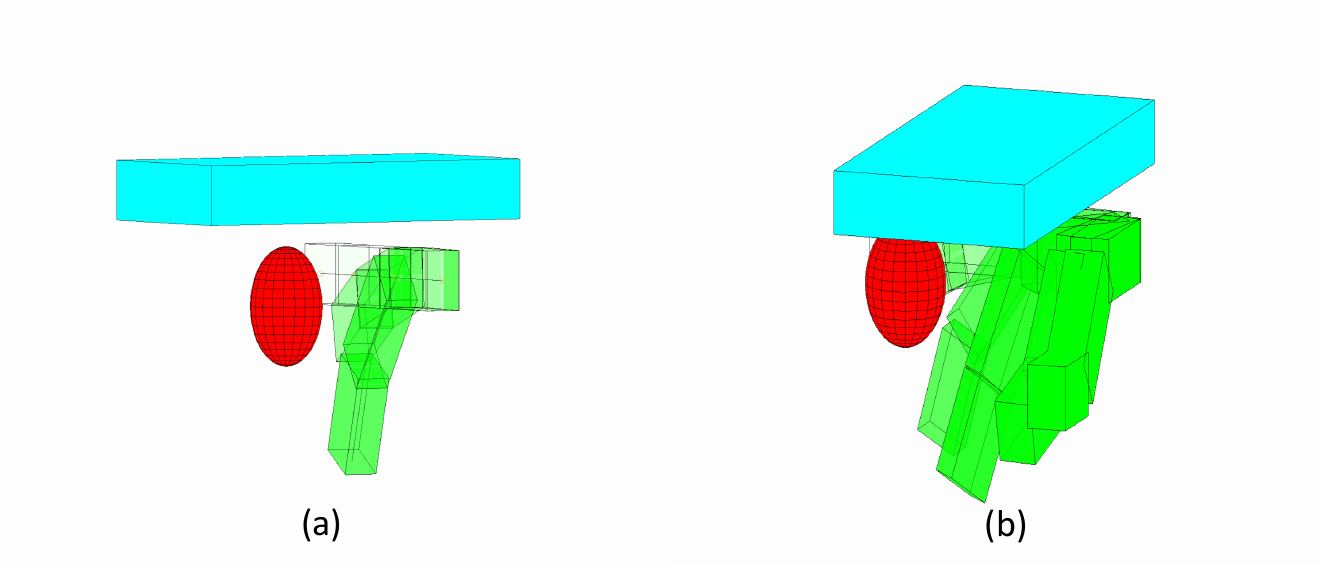}
    \caption{An example of a three-link robot avoiding static obstacles.}
    \label{fig_3link}
\end{figure}

\begin{figure}
    \centering
    \includegraphics[scale=0.6]{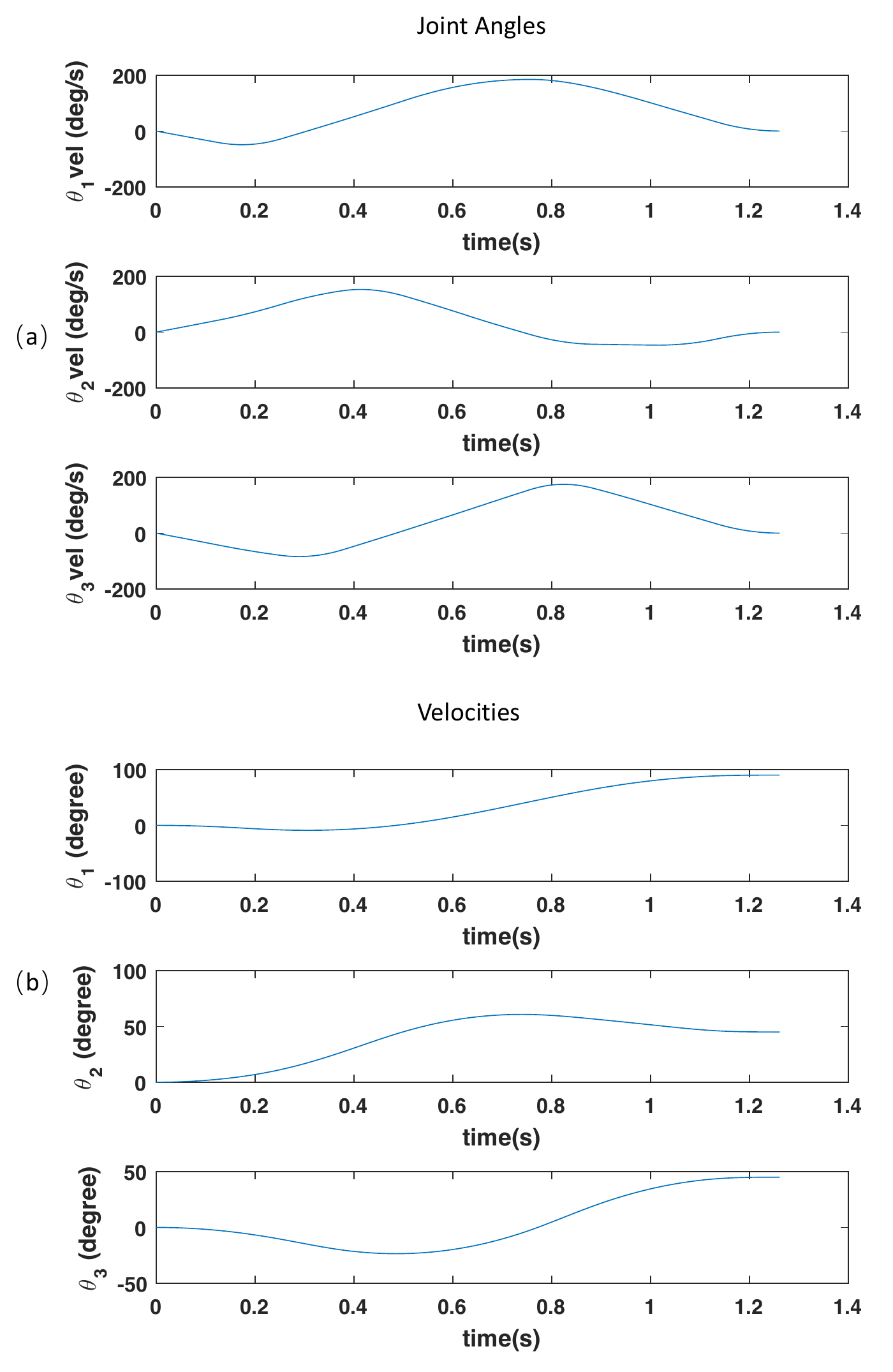}
    \caption{Joint angles and velocities profiles of the 3-link robot example.}
    \label{fig_3link_profile}
\end{figure}

Two static obstacles are placed near the robot. The objective is to minimize travel time. Joint angles, velocities, and acceleration limits are all set to $\pm 200$. Fig. \ref{fig_3link} demonstrates the movements of the robot while avoiding obstacles. Fig. \ref{fig_3link_profile} illustrates the joint angle trajectories and velocities profiles. We can see each joint trajectory is smooth and satisfies all the constraints at all time instances.


Modeling from FANUC LR Mate 200iD 7L robots, a new set of Denavit-Hartenberg parameters is used for the six-link industrial robot: 

\begin{center}
\label{table2}
\begin{tabular}{|c| c| c| c| c|} 
\hline
Link & $a_i$(m)  & $\alpha_i$(rad) & $d_i$(m) &  $\theta_i$(rad) \\ \hline
1    & $0.05$& $-\frac{\pi}{2}$ & $0$ & $\theta_1$\\ \hline
2   & $0.44$  & $\pi$ & $0$ & $\theta_2$\\ \hline
3   & $0.035$ & $-\frac{\pi}{2}$ & $0$ & $\theta_3$\\ \hline
4  & $0$ & $\frac{\pi}{2}$ & $-0.42$ & $\theta_4$\\ \hline
5  & $0$ & $-\frac{\pi}{2}$ & $0$ & $\theta_5$ \\ \hline
6  & $0$ & $\pi$ & $-0.19$ & $\theta_6$\\ \hline
\end{tabular} 
\end{center}

\begin{figure}
    \centering
    \includegraphics[scale=0.9]{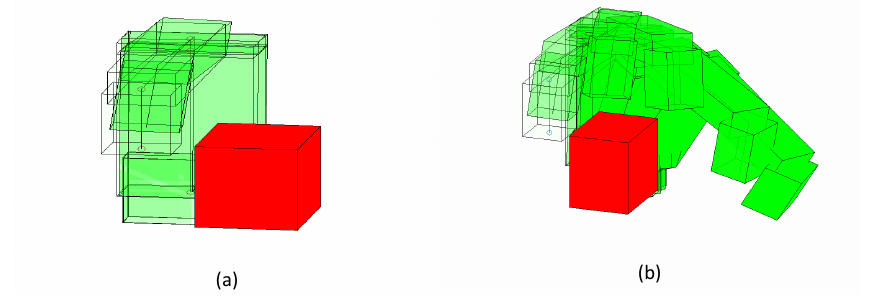}
    \caption{An example of a 6-link robot avoiding a static obstacle.}
    \label{fig_6link}
\end{figure}

\begin{figure}
    \centering
    \includegraphics[scale=0.7]{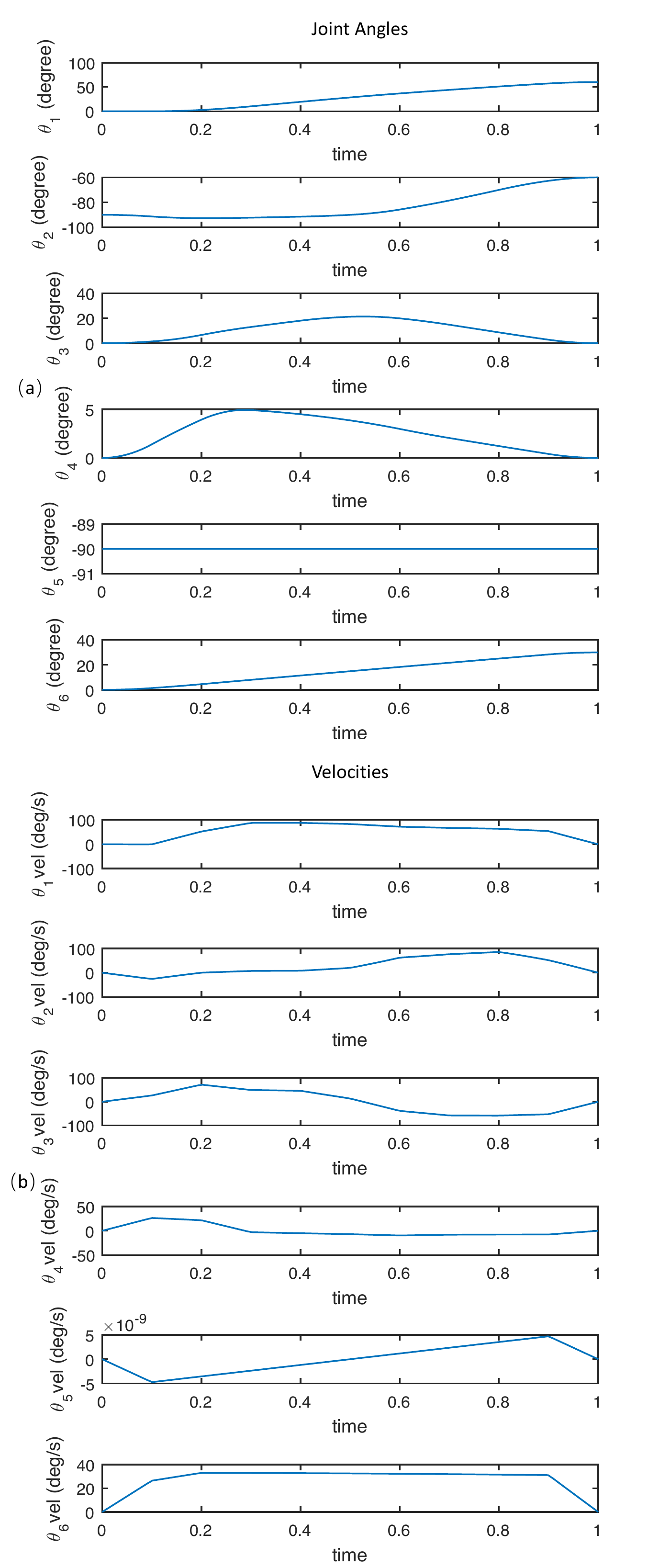}
    \caption{Joint angles and velocities profiles of the 6-link robot example.}
    \label{fig_6link_profile}
\end{figure}

Before proceeding to real experiments, a simple case was created with one static obstacle in the workspace (Fig. \ref{fig_6link}). The same objective function applies, and we set the joint angle, velocity, and acceleration limits to be $\pm 180degree$, $\pm 100 degree/s$, and $\pm 500 degree/s^2$ respectively. Fig. \ref{fig_6link_profile} illustrates the joint angle trajectories and velocity profiles. Joint angles are smooth, velocities, and accelerations (not shown) also satisfy boundary and limit constraints.


\subsection{Experimental Results on Multi-jointed Robots}
The proposed method was experimentally validated with a FANUC LR Mate 200iD 7L robot for both static and dynamic obstacles. 
Trajectories of obstacles are assumed to be known. The degree of basis functions for all B-splines $ q_i (\tau ) = \tan(\frac{\theta_i (\tau)}{2}) = \sum_{j=0}^{n} c_{ij} B_{j,3}(\tau) $ is set to $p = 3$. A union form knot vector $U = [0, 0, 0, 0, 0.1, 0.2, 0.3, \cdots,0.9, 1, 1, 1, 1]$ is used. Thus, the number of knot elements minus one results in $m = 17 - 1 = 16$, and the number of coefficients is determined by $ n+1 = m - p = 13$.
\begin{figure}
    \centering
    \includegraphics[scale=0.68]{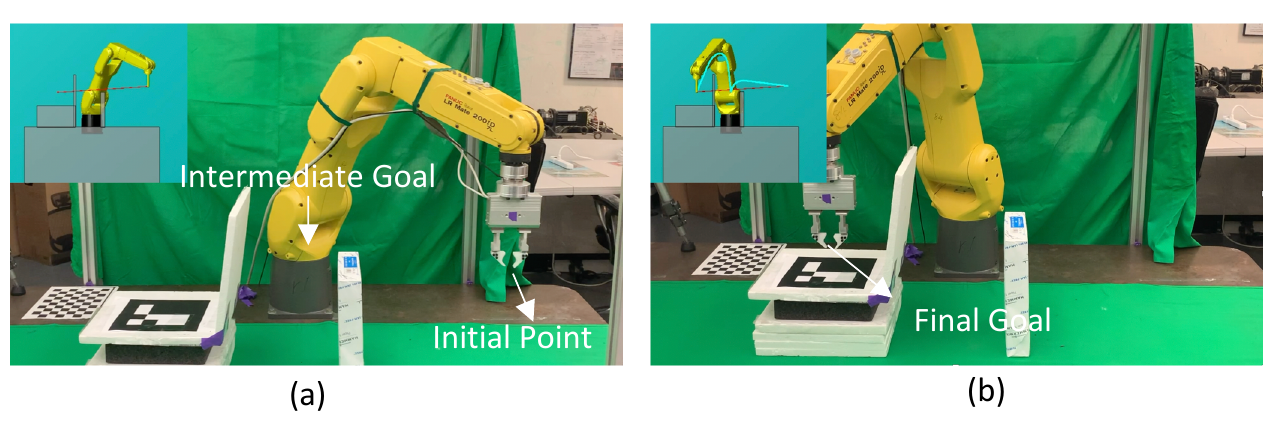}
    \caption{A FANUC robot is avoiding static thin obstacles.}
    \label{fig_exper_static}
\end{figure}

In the static case (Fig. \ref{fig_exper_static}), two obstacles were set in a row on a table. Given an initial state, the robot needs to achieve an intermediate goal between the two obstacles and then proceed to its defined final goal. The experiment shows the effectiveness of the proposed continuous formulation in the presence of thin obstacles. All constraints are satisfied for all time instances.

\begin{figure}
    \centering
    \includegraphics[scale=0.68]{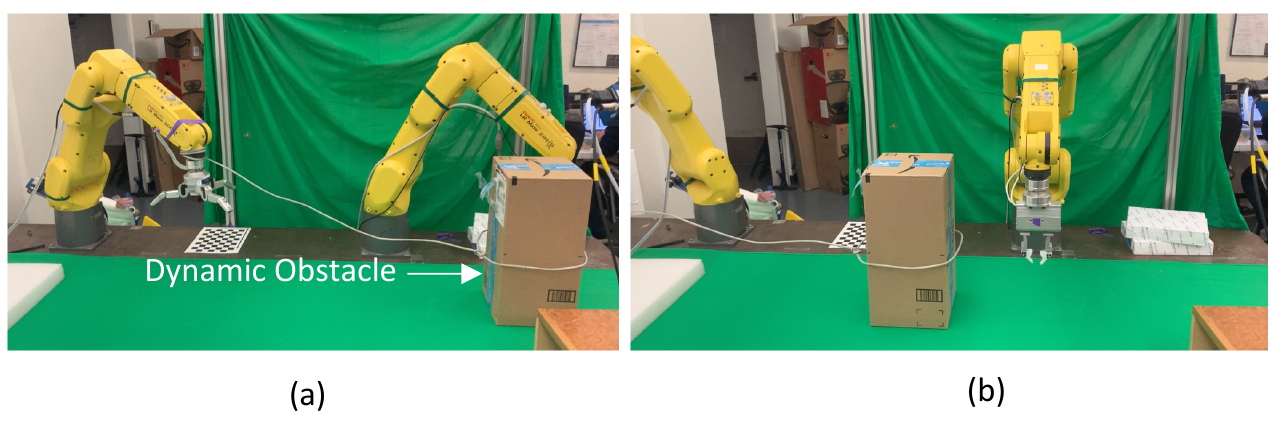}
    \caption{The experimental result on a FANUC robot avoiding a dynamic obstacle. The left robot is used to drag the obstacle to ensure it is moving at a constant velocity as a dynamic obstacle.}
    \label{fig_exper_dyn}
\end{figure}

\begin{figure}
    \centering
    \includegraphics[scale=1.3]{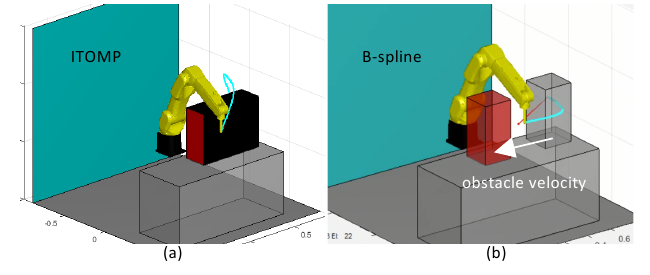}
    \caption{Comparison between the solutions of the proposed method and the ITOMP algorithm that minimize the travel distance. The dynamic obstacle is moving from the right to the left. (a) shows the simulation result of the ITOMP algorithm, where it approximates the moving obstacle with its swept volume. (b) demonstrates the result of the proposed continuous formulation, where the robot takes the advantage of trajectory timing and gets a better solution.}
    \label{fig_dynamic_compare}
\end{figure}

In the dynamic case, a second FANUC LR Mate 200iD 7L robot was used to drag the moving obstacle to ensure it is moving at a constant velocity. Given an initial position, the robot needs to navigate to its desired final position while avoiding the moving obstacle. Fig. \ref{fig_exper_dyn} shows the inital and final positions of this experiment. Again, we solved the optimization and simulated the trajectories in simulation first, then passed the trajectories to the robot for execution. A comparison was also made in this experiment between considering swept volume (an idea introduced in ITOMP~\cite{park2012itomp}) and the proposed continuous B-spline formulation. Fig. \ref{fig_dynamic_compare}(a) illustrates the end-effector trajectories in simulations for ITOMP. The swept volume formed by this moving obstacle is a long cuboid under the robot. Therefore, the robot needs to bring up its fourth link and end effector to avoid dynamic obstacles. On the other hand, the proposed method considers continuous trajectories, which allows the robot to take advantage of timing. As Fig. \ref{fig_dynamic_compare}(b) shows, the robot follows the moving obstacle to reach its end position with a much shorter trajectory. 

\section{Conclusions} 
This paper introduces a novel B-spline based trajectory optimization formulation for multi-jointed robots. 
The proposed method focuses on planning continuous collision-free trajectories by utilizing B-spline parameterization and its properties to relax optimization constraints. Kinematic and dynamic constraints are formulated via rigorously defined B-spline operations. The method utilizes signed distance fields and separating hyperplanes to prevent collisions with static and dynamic obstacles respectively. The use of signed distance fields contributes to shorter computation time in static environments as the number of obstacles increases, while constructing separating hyperplanes is able to consider dynamic obstacles naturally with time.
Simulation results show the effectiveness of the method under various scenarios with different kinds of robots. Experimental validation is done using a 6-link FANUC robot to plan and execute continuous collision-free trajectories with static and moving obstacles. 



\bibliographystyle{IEEEtran}
\bibliography{references}	
\vspace{12pt}
\end{document}